\documentclass[10pt,twocolumn,letterpaper]{article}

\usepackage{cvpr}              

\usepackage{graphicx}
\usepackage{amsmath}
\usepackage{amssymb}
\usepackage{amsthm}
\usepackage{amsfonts}
\usepackage{mathtools}
\usepackage{booktabs}
\usepackage[normalem]{ulem}
\usepackage{enumerate}
\usepackage{enumitem}

\usepackage[pagebackref,breaklinks,colorlinks]{hyperref}

\usepackage[capitalize]{cleveref}
\crefname{section}{Sec.}{Secs.}
\Crefname{section}{Section}{Sections}
\Crefname{table}{Table}{Tables}
\crefname{table}{Tab.}{Tabs.}

\def\cvprPaperID{*****} 
\def\confName{CVPR}
\def\confYear{2023}

\newcommand{\sv}[1]{{{#1}}}
\newcommand{\sa}[1]{{{#1}}}
\newcommand{\lp}[1]{{{#1}}}

\begin{document}

\title{Reassembling Broken Objects using Breaking Curves}

\author{
\begin{tabular}{c@{\hskip 0.8cm}c@{\hskip 0.8cm}c@{\hskip 0.8cm}c@{\hskip 0.25cm}c}
    Ali Alagrami\thanks{Equal contribution. } & Luca Palmieri\footnotemark[1] & Sinem Aslan\footnote[1] & Marcello Pelillo & Sebastiano Vascon\\
    \multicolumn{5}{c}{{DAIS, Ca’ Foscari University of Venice, Italy}} \\
    \multicolumn{5}{c}{\texttt{\small887727@stud.unive.it, \{luca.palmieri, sinem.aslan, pelillo,  sebastiano.vascon\}@unive.it}} \\
  \end{tabular}
}
\maketitle

\begin{abstract}
Reassembling 3D broken objects is a challenging task. A robust solution that generalizes well must deal with diverse patterns associated with different types of broken objects. We propose a method that tackles the pairwise assembly of 3D point clouds, that is agnostic on the type of object, and that relies solely on their geometrical information, without any prior information on the shape of the reconstructed object. The method receives two point clouds as input and segments them into regions using detected closed boundary contours, known as \emph{breaking curves}. Possible alignment combinations of the regions of each broken object are evaluated and the best one is selected as the final alignment. Experiments were carried out both on available 3D scanned objects and on a recent benchmark for synthetic broken objects. Results show that our solution performs well in reassembling different kinds of broken objects.
\end{abstract}

\section{Introduction}

Reconstructing three-dimensional broken objects is an important task in several fields such as computer graphics \cite{Huang2006, breaking_bad}, cultural heritage \cite{survey_ch, Papaioannou2017}, and robotics \cite{kataoka2022bi, ghasemipour2022blocks, yu2021roboassembly}.
The growing interest in the community toward the 3D multi-part assembly task in recent years led to the development of \lp{a benchmark composed of realistically broken objects}\cite{breaking_bad}.

While there are numerous methods for the registration of 3D points, e.g., \cite{Besl1992AMF, zhang1994iterative, teaser, predator}, reassembling two parts of a broken object is a different task that usually requires registering only a partial subset of each part.
Some registration methods address this issue by focusing on the low-overlap region \cite{predator}, however accurately identifying the fractured surface region is important for performing pairwise matching over such point subsets. Indeed, the success of the reassembly depends highly on the precision of the segmentation process, and developing an algorithm that accurately identify fractured surface regions without making assumptions about the shape of the object is challenging. To deal with this issue, prior works \cite{altantsetseg2014pairwise, Huang2006, son2018reassembly} adopted extraction of \textit{breaking curves} in an initial step, and achieve segmentation by merging vertices that are not part of the breaking curve into a single region.
\lp{Other approaches adopted graph-based techniques for segmentation of point clouds, as outlined in Section \ref{sec:2}.
These have been successfully used for extracting spatial geometric attributes from 3D point cloud data \cite{hao2022mixed, loizou2020learning, Gumhold2001}.}

\begin{figure}[t]
    \begin{subfigure}{.1\textwidth}
      \centering
      \includegraphics[width=.9\linewidth]{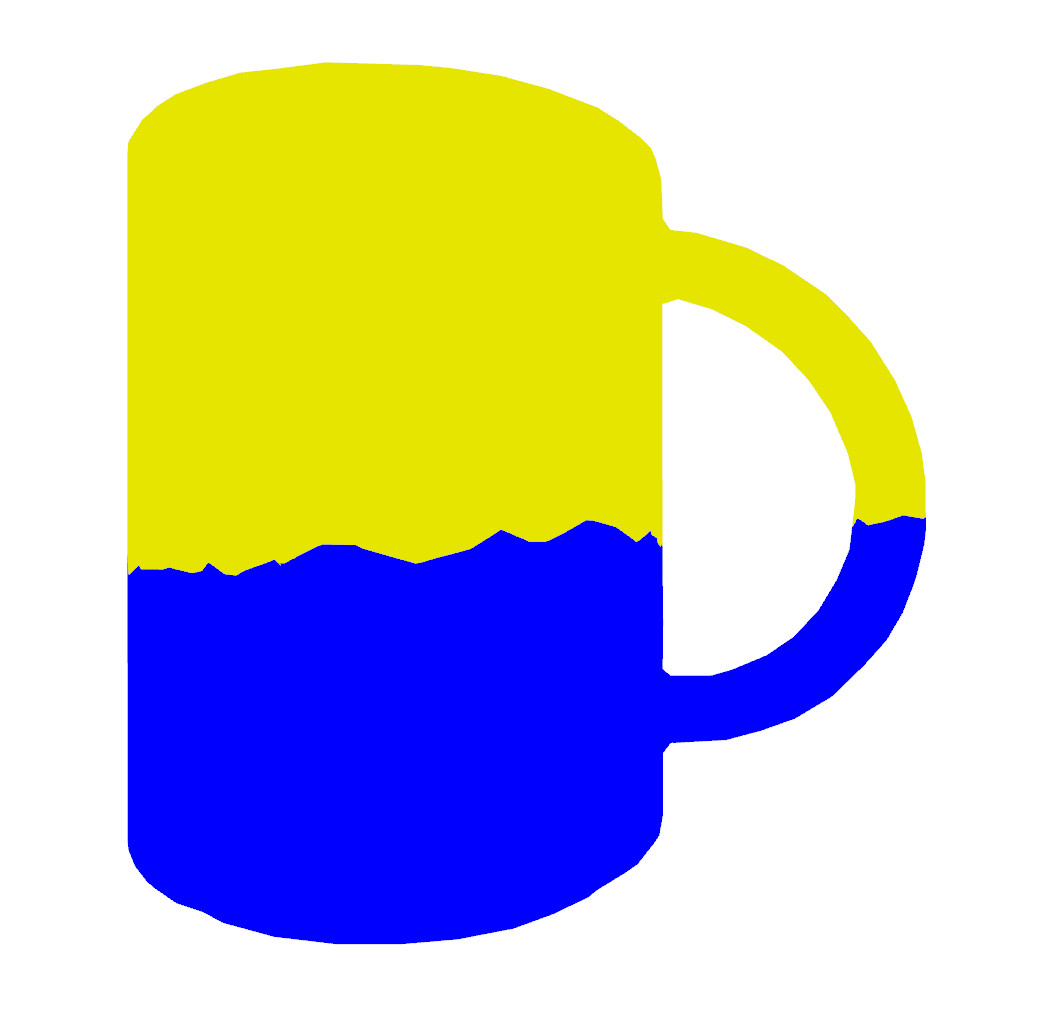}
      \caption{Original}
      \label{fig:sfig1}
    \end{subfigure}%
    \begin{subfigure}{.1\textwidth}
      \centering
      \includegraphics[width=.9\linewidth]{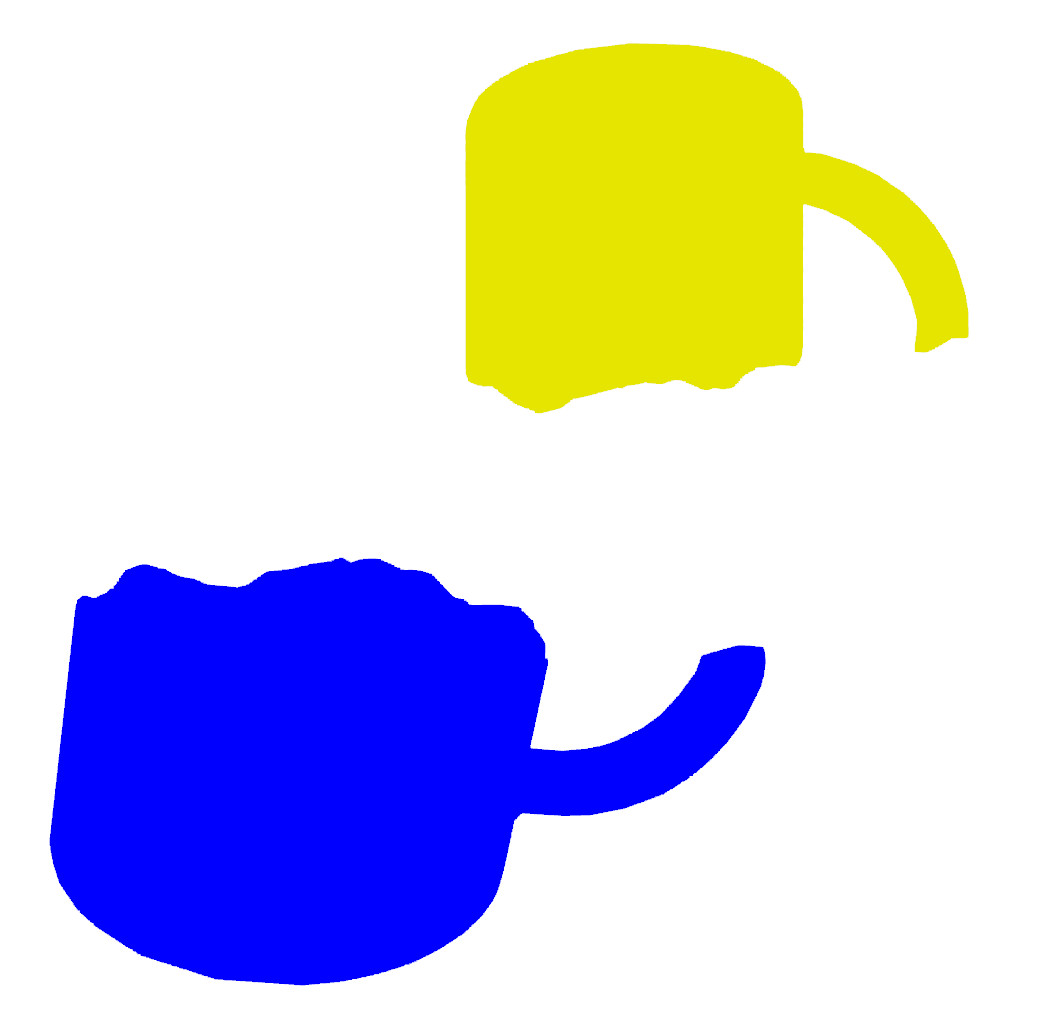}
      \caption{Input}
      \label{fig:sfig1}
    \end{subfigure}%
    \begin{subfigure}{.1\textwidth}
      \centering
      \includegraphics[width=.9\linewidth]{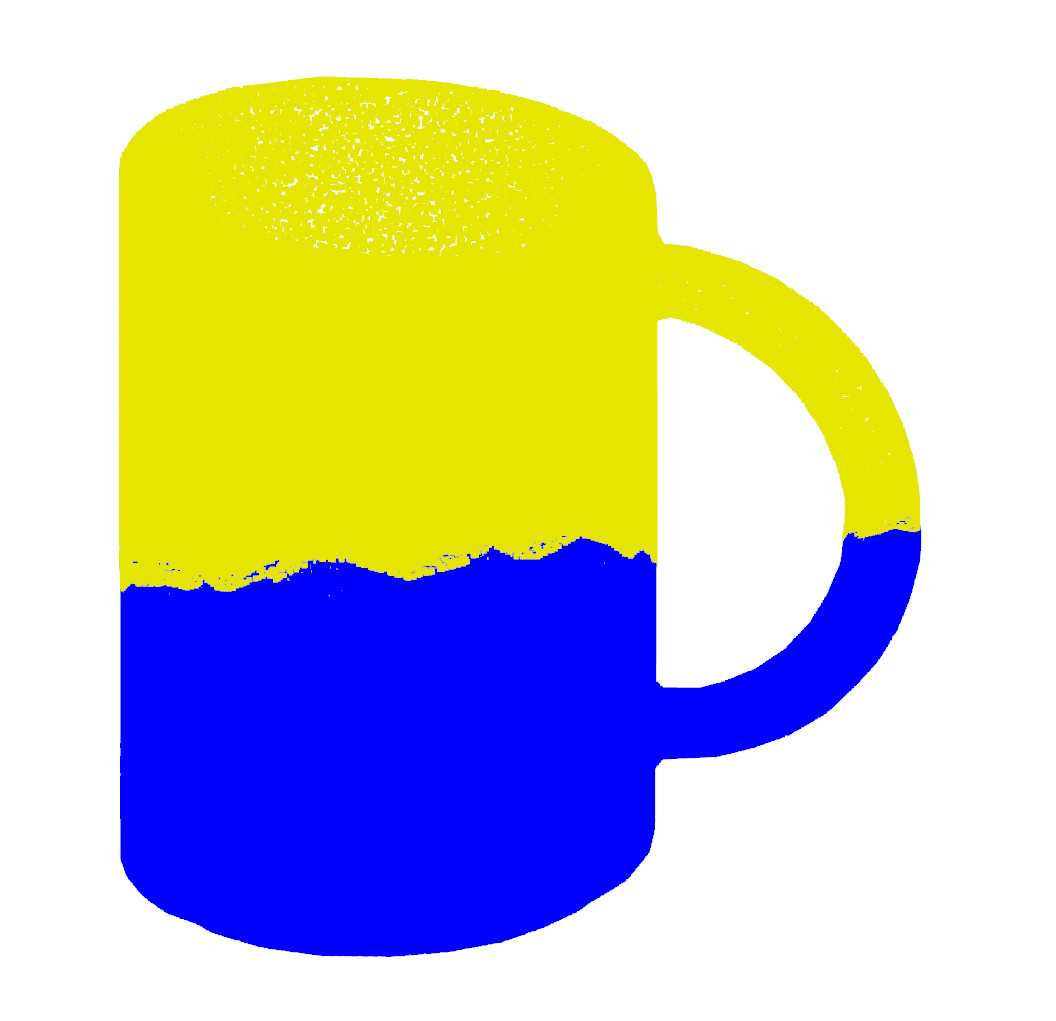}
      \caption{Proposed}
      \label{fig:sfig1}
    \end{subfigure}%
    \begin{subfigure}{.1\textwidth}
      \centering
      \includegraphics[width=.9\linewidth]{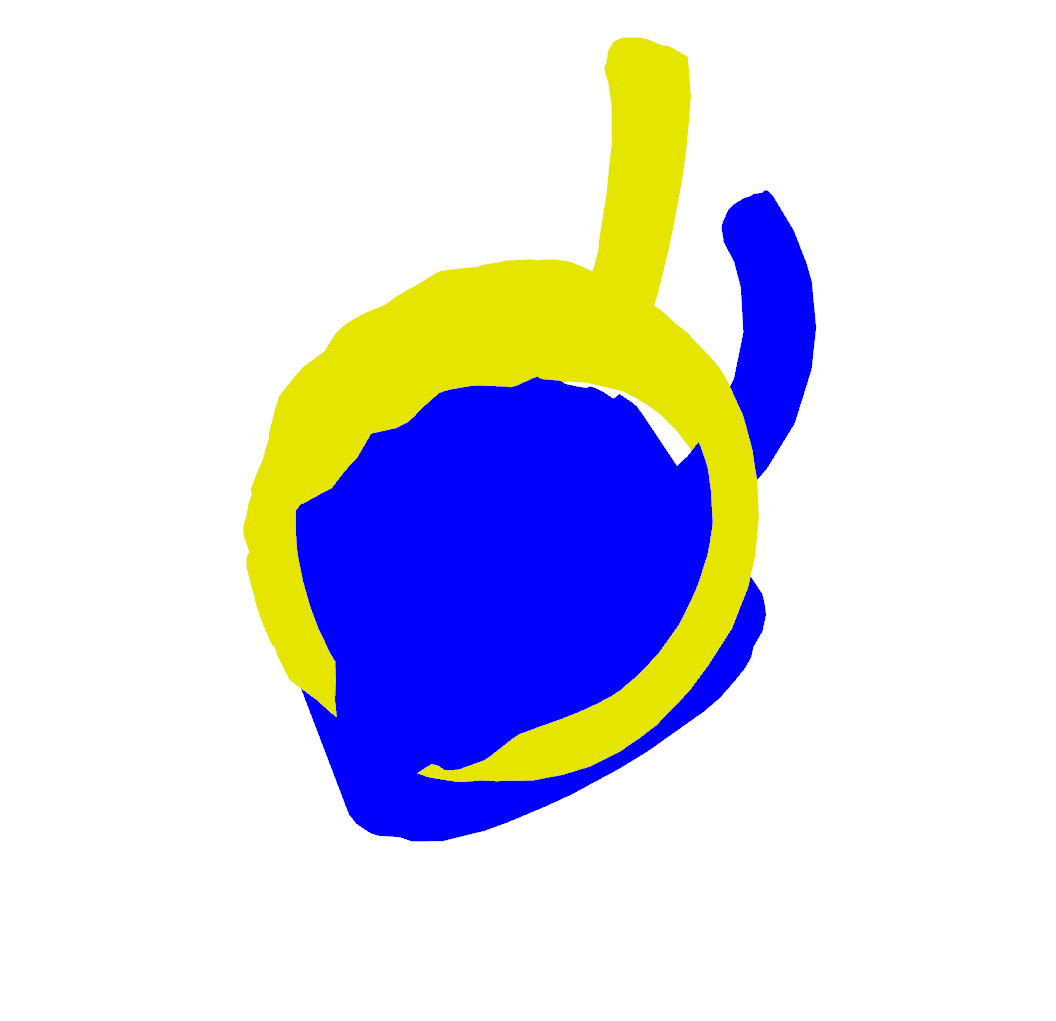}
      \caption{DGL \cite{HuangZhan2020PartAssembly}}
      \label{fig:sfig1}
    \end{subfigure}%
    \begin{subfigure}{.1\textwidth}
      \centering
      \includegraphics[width=.9\linewidth]{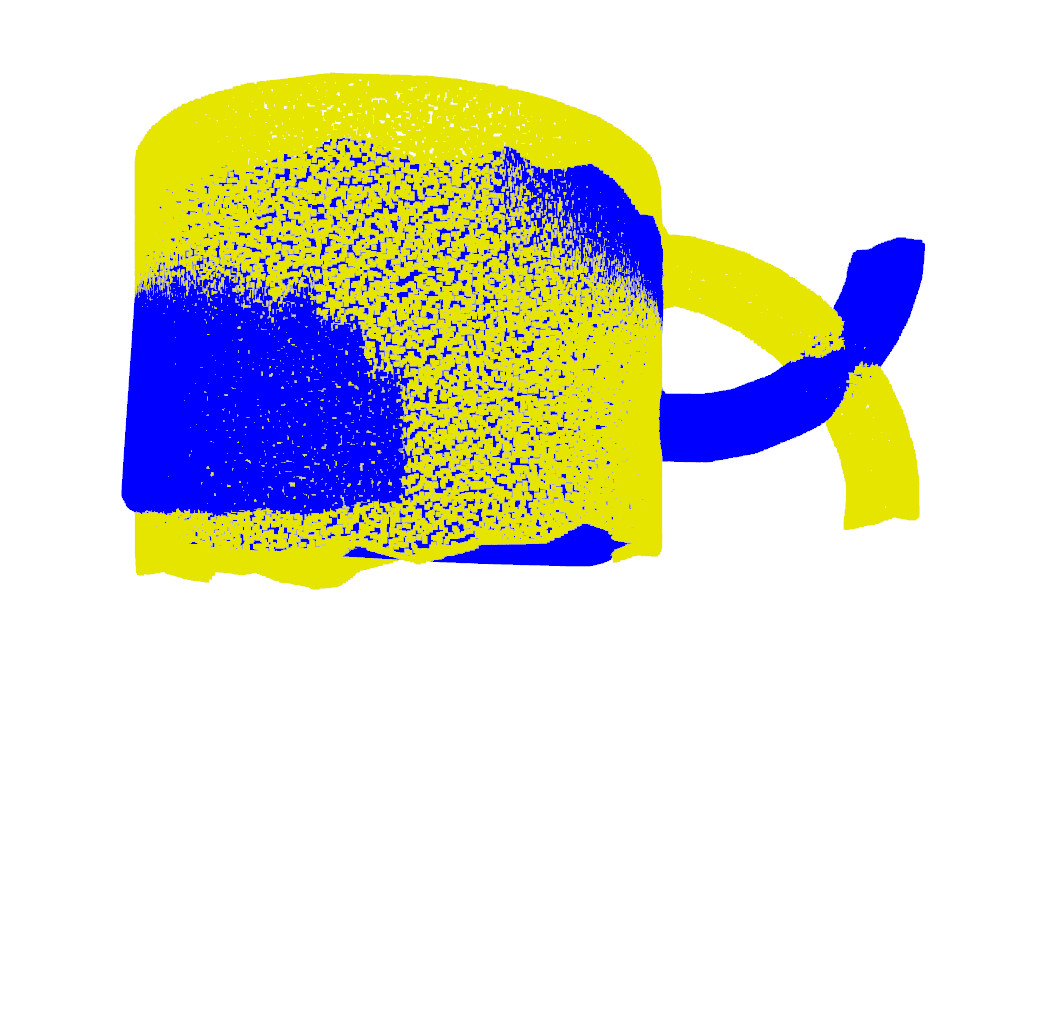}
      \caption{ICP \cite{Besl1992AMF}}
      \label{fig:sfig1}
    \end{subfigure}%
    \caption{\footnotesize The proposed method reconstructs accurately the mug by assembling the two parts, where the other approaches fail drastically in this case.}
    \label{fig:results}
    \vspace{-0.3cm}
\end{figure}

We propose a modular and adaptable open-source\footnote{The code will be released in \href{https://github.com/RePAIRProject/AAFR}{https://github.com/RePAIRProject/AAFR}.} framework that integrates geometric-based methods to effectively reassemble pairs of 3D broken objects, without making any assumptions about their type or the nature of their damage.
The proposed approach offers a significant advantage in obtaining region segmentation independent of surface characteristics. This is achieved through the guidance of \emph{breaking curves}, which are extracted using an extension of the graph-based method in \cite{Gumhold2001}.
We experimentally demonstrate that, if the breaking curve extraction and the successive segmentation steps are successfully achieved, it is possible to accomplish the registration stage with a standard registration method such as the Iterative Closest Point (ICP) \cite{Besl1992AMF}.
We evaluated the proposed approach on a state-of-the-art synthetic benchmark as well as two real-world datasets.
The results demonstrate the robustness and accuracy of the proposed method, as presented in Figure \ref{fig:results}.

\begin{figure*}[t]
    \begin{subfigure}{.14\textwidth}
      \centering
      \includegraphics[height=1.8\textwidth]{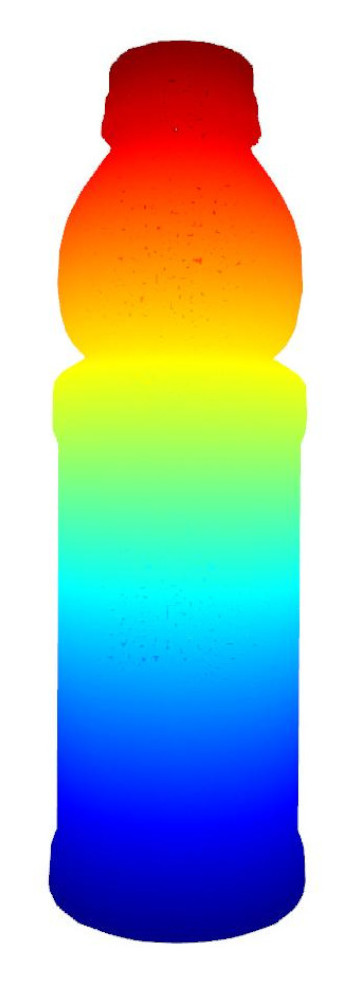}
      \caption{DrinkBottle}
      \label{fig:sfig1}
    \end{subfigure}%
    \begin{subfigure}{.14\textwidth}
      \centering
      \includegraphics[height=1.8\textwidth]{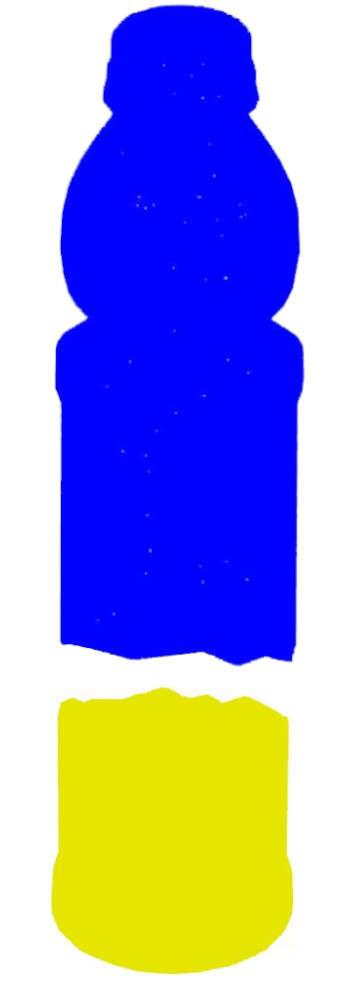}
      \caption{Broken Parts}
      \label{fig:sfig1}
    \end{subfigure}%
    \begin{subfigure}{.1\textwidth}
      \centering
        \includegraphics[width=\textwidth]{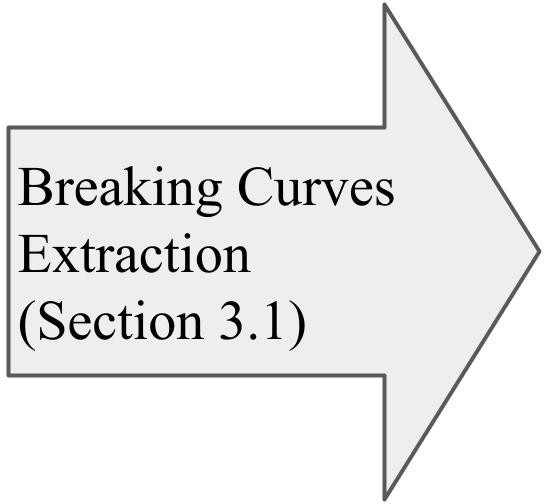}
        \vspace*{4em}

      \label{fig:sfig1}
    \end{subfigure}%
    \begin{subfigure}{.14\textwidth}
      \centering
      \includegraphics[height=1.8\textwidth]{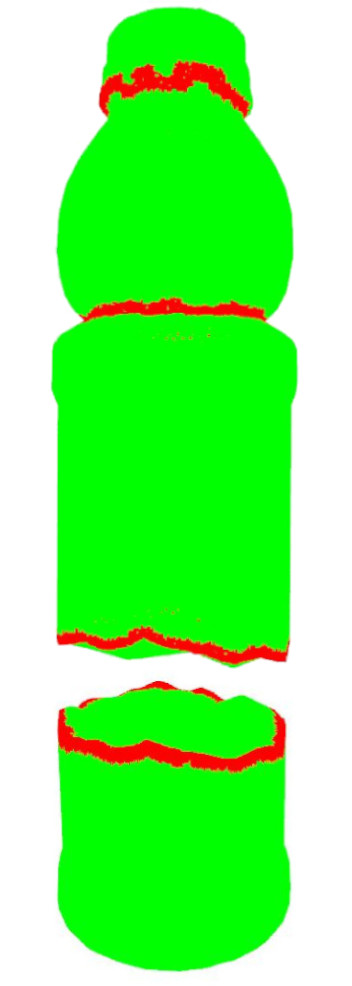}
      \caption{Breaking curves}
      \label{fig:sfig1}
    \end{subfigure}%
    \begin{subfigure}{.1\textwidth}
        \centering
        \includegraphics[width=\textwidth]{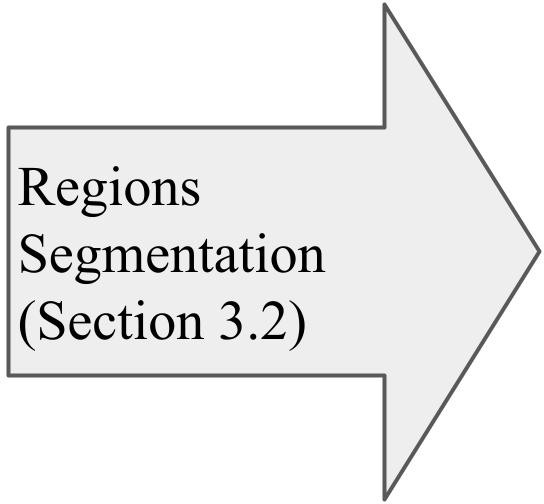}
        \vspace*{4em}

      \label{fig:sfig1}
    \end{subfigure}%
    \begin{subfigure}{.14\textwidth}
      \centering
      \includegraphics[height=1.8\textwidth]{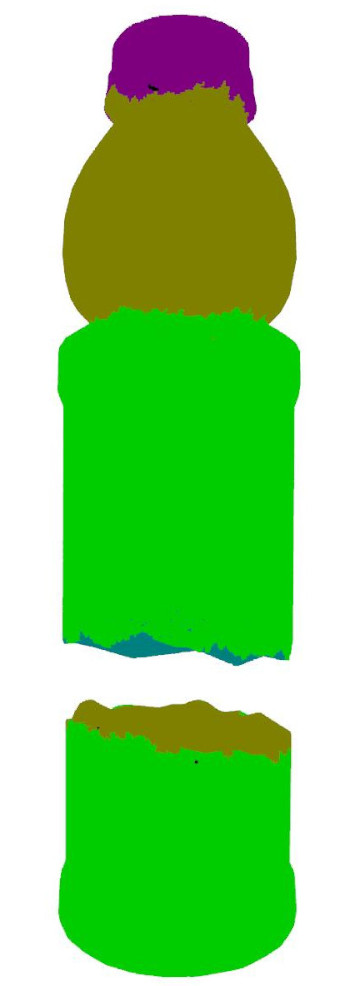}
      \caption{Regions}
      \label{fig:sfig1}
    \end{subfigure}%
    \begin{subfigure}{.1\textwidth}
         \centering
        \includegraphics[width=\textwidth]{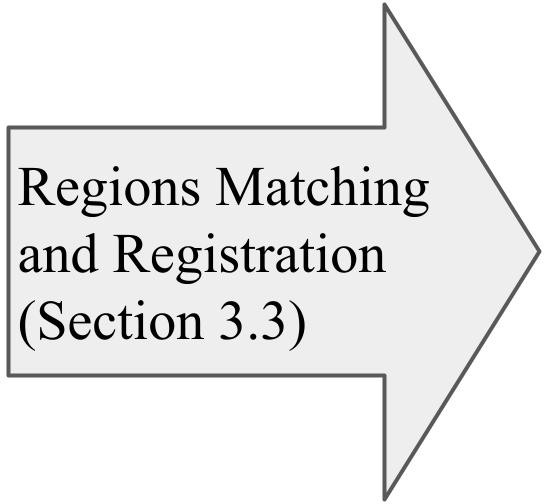}
        \vspace*{4em}

      \label{fig:sfig1}
    \end{subfigure}%
    \begin{subfigure}{.14\textwidth}
      \centering
      \includegraphics[height=1.8\textwidth]{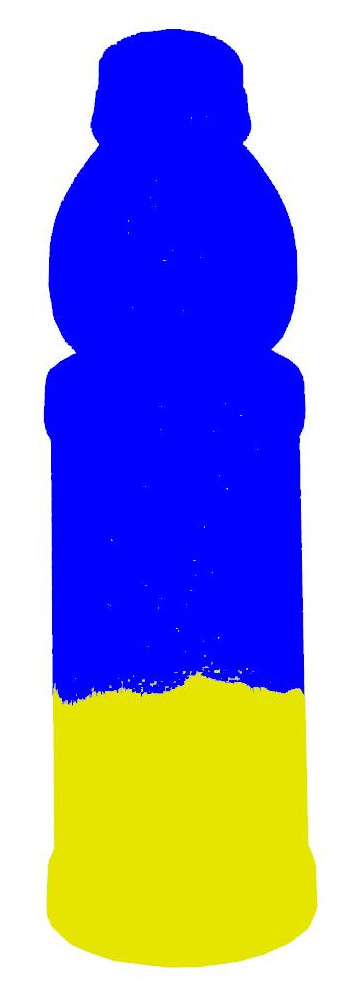}
      \caption{Reassembled}
      \label{fig:sfig1}
    \end{subfigure}%
    \caption{\footnotesize The pipeline of the proposed approach. Two broken parts of a 3D point cloud (a bottle) are the input to the algorithm. After processing, the breaking curves are extracted and the point clouds are segmented. The registration selects the best match among the segmented parts and aligns the two input point clouds. The point cloud belongs to the DrinkBottle category of the breaking bad dataset \cite{breaking_bad}.}
    \label{fig:pipeline}
\end{figure*}

\section{Related Work}
\label{sec:2}

\textbf{Non-learning based (geometrical) methods:} A common approach for automatic reassembly of broken 3D objects relies on fractured region matching for identifying potential pairwise matches of fragments. This involves \textit{(i)} segmentation of the broken objects into fractured and intact regions and (\textit{ii)} matching of the fractured surfaces. A conventional technique for surface segmentation is to use \textit{region growing}, where vertices with similar attributes are combined in the same region.

The region growing segmentation relies either on the contours or on the surface characteristics.
Altantsetseg et al. \cite{altantsetseg2014pairwise} adopted the Fourier series to approximate the boundary contour, Huang et al. \cite{Huang2006} extracted the long closed cycles from a minimum spanning graph of the edge points that have persistent curvatures at multiple scales.
Several works used breaking curves for aligning fragments after segmenting them \cite{yang2017pairwise, zhao2018rigid}, yet they do not consider deteriorated fragments.
Some other works adopted features computed on the fractured surfaces for their alignment, e.g., concave and convex regions were extracted on the fractured surfaces by Li et al. \cite{li2019pairwise} and Son et al. \cite{son2018reassembly}, and Huang et al. \cite{Huang2006} adopted clusters of multi-scale surface characteristics computed based on the integral invariants.
Papaioannou et al. \cite{Papaioannou2017} conducted an exhaustive search of fractured surfaces of all fragments, rather than extracting features.

\textbf{Learning-based methods:} Another approach adopted by the recent literature involves learning-based techniques to estimate the transformation required for the reassembly of fragments. In this context, Chen et al. \cite{chen2022neural} created a synthetic dataset by breaking 3D meshes into pairs of fragments
and employed a transformer-based network with a loss that is a combination of geometric shape-based and transformation matrix-based loss functions to learn pairwise alignment.
The reported results highlight the high complexity of this task, given that synthetically generated fragments devoid of physical deterioration were only roughly aligned \cite{breaking_bad}.
This trend is further validated by Sellan et al. \cite{breaking_bad}, which introduced a physically realistic dataset of broken 3D meshes to serve as a benchmark for the reassembly task and demonstrated that baseline learning-based algorithms are insufficient for solving the multi-part assembly task.

In this work, we follow the first approach, i.e., segment the broken surfaces as in \cite{altantsetseg2014pairwise, Huang2006, son2018reassembly} and register each segmented broken region with an exhaustive search as in \cite{Papaioannou2017}.
Unlike them, we use a graph-based method for detecting the 
breaking curves of fragments which allows segmenting regions without prior assumptions on the surface characteristics of the object, and adopt the ICP algorithm for registration.

\section{The Proposed Approach}
The proposed method has a modular workflow depicted in Figure \ref{fig:pipeline}, which is divided into three main parts:

\begin{enumerate}[itemsep=-0.3em]
    \item Detecting breaking curves: the set of points which belong to a three-dimensional edge (Section \ref{subs:br_curves}),
    \item Segmenting the points into a set of regions using the breaking curves (Section \ref{subs:segmentation}),
    \item Registering the objects by selecting the best match among possible combinations of the segmented regions of each objects (Section \ref{subs:registration}).
\end{enumerate}

\subsection{Breaking Curves Extraction}
\label{subs:br_curves}
\begin{figure}[h!]
    \begin{subfigure}{.16\textwidth}
      \centering
      \includegraphics[width=.85\linewidth]{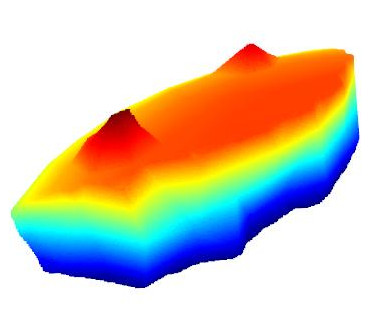}
      \includegraphics[width=.81\linewidth]{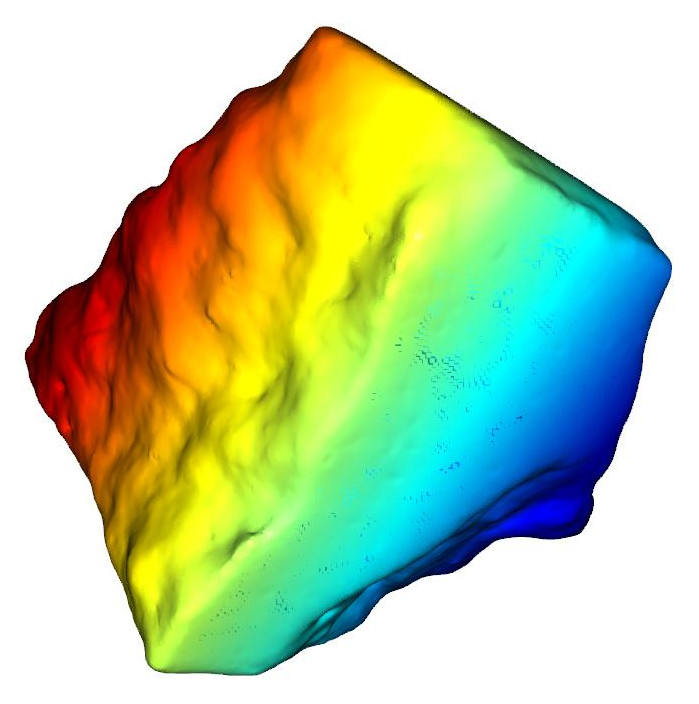}
      \caption{Original}
      \label{fig:sfig1}
    \end{subfigure}%
    \begin{subfigure}{.16\textwidth}
      \centering
      \includegraphics[width=.85\linewidth]{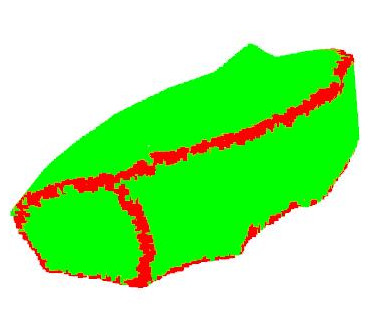}
      \includegraphics[width=.81\linewidth]{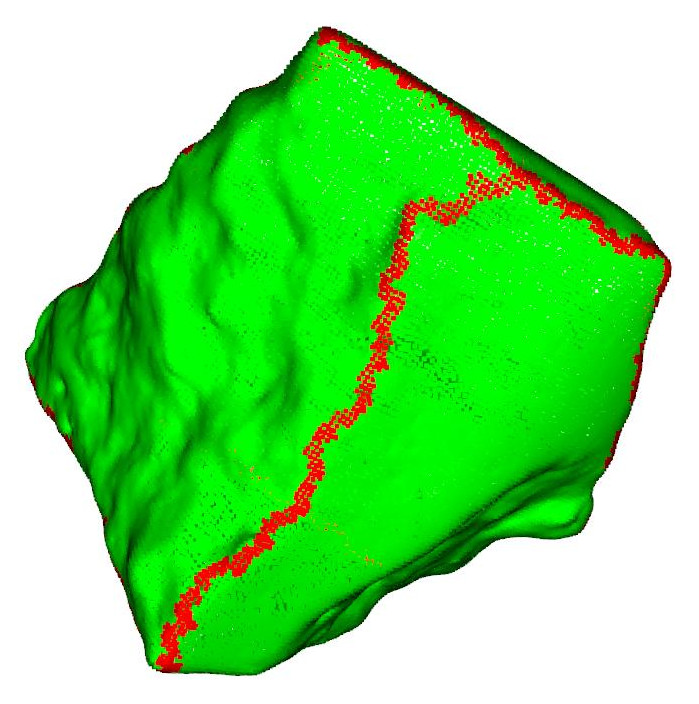}
      \caption{Breaking Curves}
      \label{fig:subfig:br_curves}
    \end{subfigure}%
    \begin{subfigure}{.16\textwidth}
      \centering
      \includegraphics[width=.85\linewidth]{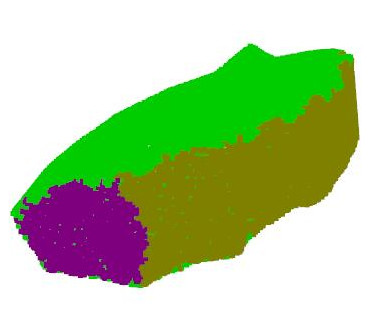}
      \includegraphics[width=.81\linewidth]{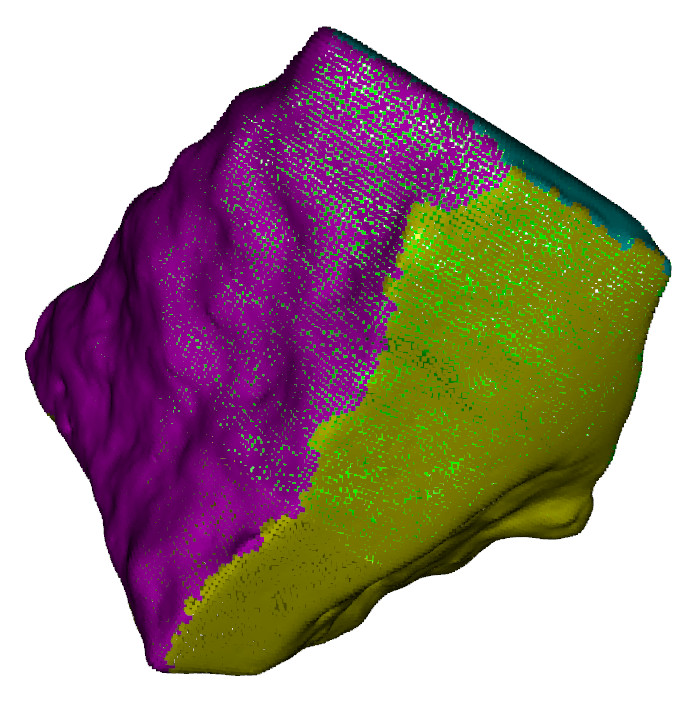}
      \caption{Segmented Regions}
      \label{fig:sfig1}
    \end{subfigure}%
    \caption{An example of the pipeline on both synthetic (top) and real (bottom) data: after processing the original point cloud, the borders (in red) are detected and the regions are segmented accordingly (different colors).}
    \label{fig:segmentation}
    \vspace{-0.3cm}
\end{figure}

When dealing with the assembly of fragmented objects, it is crucial to detect borders and edges as they provide cues for the correct matching. The proposed approach starts from a 3D point cloud and detects breaking curves. A breaking curve is defined as a subset of connected points that belong to a 3D edge, as illustrated in Figure \ref{fig:subfig:br_curves}. The set of all breaking curves acts as a support for segmenting the objects into distinct regions.

\sv{Let $P$ be the set of points in a point cloud. We represent $P$ as an unweighted directed graph $G=(V,E)$ where the set of vertices $V$ corresponds to the set of points $p \in P$ and the edges $E\subseteq V \times V$ represents the neighbouring relations between the points.
Being the density of the point cloud non-uniform, we opted for a mixed approach when adding edges:}
\lp{we create an $\epsilon$-graph \cite{natali2011graph,yu2018ec} using the average distance of the $k$ nearest neighbours considering the entire point cloud. The $\epsilon$ value is then computed as:}

\begin{equation}
\nonumber
\epsilon = \frac{1}{|P|}{\frac{1}{k}\sum_{p \in P}\sum_{q \in {\mathcal{N}_p^k}} |p-q|}
\end{equation}
\sv{Here $P$ is the point cloud, $p \in P$ is a 3D-point in $\mathbb{R}^3$ and $\mathcal{N}_p^k$ is the set of $k$-nearest neighbours of point $p$.}

\sv{After the graph is created, we compute for each node its \emph{corner penalty} \cite{Gumhold2001} defined as:}
\begin{equation}
\footnotesize
\nonumber
\omega_{co}(p) = \dfrac{\lambda_2(p)-\lambda_0(p)}{\lambda_2(p)}
\label{w_co}
\end{equation}
\sv{where $\lambda_0$ and $\lambda_2$ are respectively the smallest and the largest of the three eigenvalues of the correlation matrix of the neighbours of $p$. The eigenvalues of the correlation matrix provide the level of skewness of the ellipsoid enclosing the points.}
Intuitively, if the point $p$ lies on a flat area (i.e. the surface), one would have $\lambda_2\approx\lambda_1$ and $\lambda_2\approx0$, while if the point lies on a corner, the eigenvalues should approximately be the same $(\lambda_2\approx\lambda_1\approx\lambda_0)$ \cite{Gumhold2001}. If the corner penalty tends to $1$, the node is likely to be on a flat area.
\lp{We select all nodes whose corner penalty is less than a threshold to obtain a noisy initial version of the \emph{breaking curves}.
The final version is obtained after applying a refinement step similar to the morphological operation of opening. A pruning step is followed by a dilation to remove small isolated branches and promote the creation of closed breaking curves.} \sv{Given a point cloud $P$ we define $\mathcal{B}^P$ as the set of points in $P$ that are part of a breaking curve.}

\subsection{Regions Segmentation}
\label{subs:segmentation}

\sv{Regions are extracted using a region-growing approach constrained by the previously extracted breaking curves. Given a point $p \notin \mathcal{B}^P$ we define the $i$-th region $\mathcal{R}_i^P$ and assign $p$ to it. We consider the set of $q \in \mathcal{N}_p$ and include each $q$ in the region $\mathcal{R}_i$ if $q \notin \mathcal{B}^P$. This procedure is iterated until all $p \notin \mathcal{B}^P$ are considered. This results in segmenting the point cloud $P$ into several regions $\mathcal{R}^P$ enclosed by the breaking curves.}

The only points that remain unassigned to a region are those that belong to the breaking curves.
However, the breaking curve shape can also aid in the matching phase.
Thus, a $k$-NN voting scheme is employed to assign these points to a segmented region, i.e., if the majority of neighboring points of a breaking curve point belong to a particular region, then it is assigned to that region.

\subsection{Region Matching and Registration}
\label{subs:registration}

The final step involves aligning the fragments using the segmented regions. Given two segmented point clouds $P$ and $Q$, we attempt to register the regions in $\mathcal{R}^P$ with the one in $\mathcal{R}^Q$. To this end, we first discard regions having a number of nodes below a certain threshold. This step has two beneficial effects: reducing the computational effort and making the method more robust to noisy regions. The registration is achieved with an exhaustive search of all the remaining regions matches. Given a pair of regions $\mathcal{R}_i^P$ and $\mathcal{R}_j^Q$, we register them with ICP \cite{Besl1992AMF} and compute the Chamfer Distance (CD) as their matching score. The pair with the best score is selected and their transformation is used for the final alignment.

\section{Experiments}
We reported results of our model considering two available datasets of both synthetic and real scanned 3D objects and an in-house set of scanned 3D fresco fragments from the Pompeii Archaeological Site collected under the RePAIR project\footnote{For more information, please visit \href{https://www.repairproject.eu/}{https://www.repairproject.eu/}.}. \sa{In particular, we experimented on a subset of categories of the Breaking Bad (BBad) dataset \cite{breaking_bad} having enough variability in terms of object characteristics, and one sample of TU-Wien dataset \cite{Huang2006} since it was sufficient to explore} whether the proposed algorithm is capable of solving the reassembly task for different objects.

\sa{We compare our method} against the Generative 3D Part Assembly (DGL) method proposed in \cite{HuangZhan2020PartAssembly}, which was \sa{reported as the superior method} on the BBad dataset in \cite{breaking_bad}.
As a baseline we also include ICP \cite{Besl1992AMF} into our evaluation
\footnote{We trained from scratch the DGL on only pairs of fragments following \href{https://github.com/Wuziyi616/multi_part_assembly/blob/master/docs/model.md}{authors' implementation} and used the \href{http://www.open3d.org/docs/release/tutorial/pipelines/icp_registration.html}{Open3D implementation} for ICP.}.

\begin{figure*}[t!]
    \begin{subfigure}{.125\textwidth}
      \centering
      \includegraphics[width=.9\linewidth]{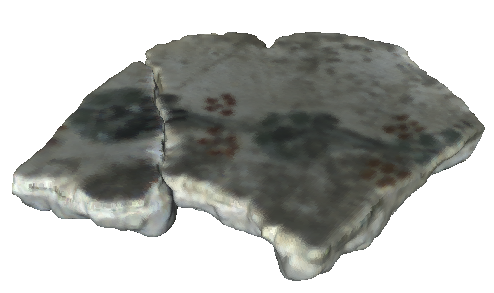}
      \caption{\textit{RePAIR}}
      \label{fig:sfig1}
    \end{subfigure}%
    \begin{subfigure}{.125\textwidth}
      \centering
      \includegraphics[width=.9\linewidth]{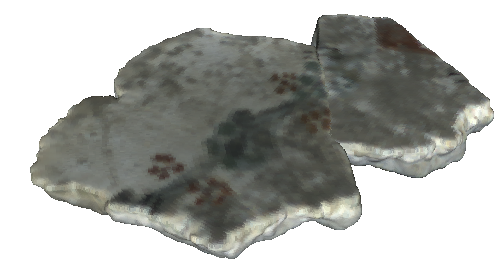}
      \caption{\textit{RePAIR}}
      \label{fig:sfig1}
    \end{subfigure}%
    \begin{subfigure}{.125\textwidth}
      \centering
      \includegraphics[width=.9\linewidth]{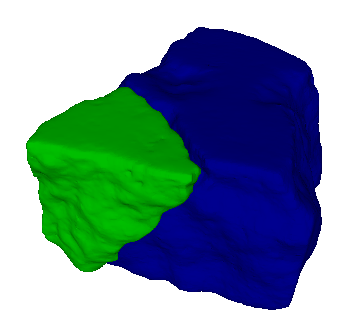}
      \caption{Brick \cite{Huang2006}}
      \label{fig:sfig1}
    \end{subfigure}%
    \begin{subfigure}{.125\textwidth}
      \centering
      \includegraphics[width=.9\linewidth]{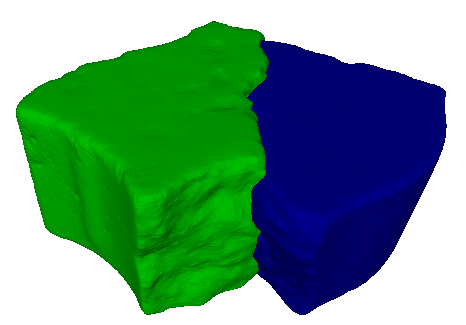}
      \caption{Brick \cite{Huang2006}}
      \label{fig:sfig1}
    \end{subfigure}
    \begin{subfigure}{.125\textwidth}
      \centering
      \includegraphics[width=.72\linewidth]{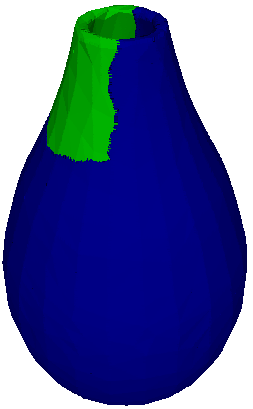}
      \caption{Vase\cite{breaking_bad}}
      \label{fig:sfig1}
    \end{subfigure}%
    \begin{subfigure}{.125\textwidth}
      \centering
      \includegraphics[width=.75\linewidth]{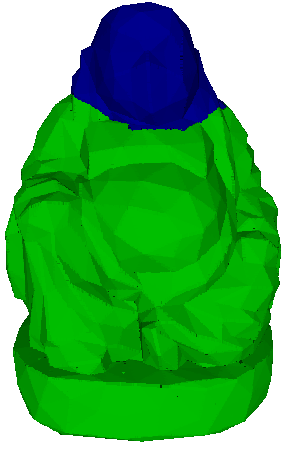}
      \caption{Statue\cite{breaking_bad}}
      \label{fig:sfig1}
    \end{subfigure}%
    \begin{subfigure}{.125\textwidth}
      \centering
      \includegraphics[width=.735\linewidth]{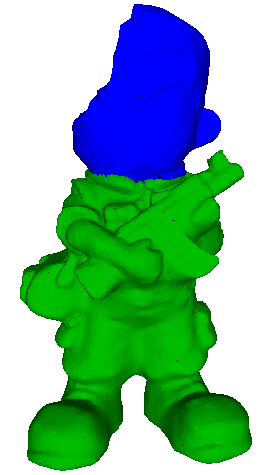}
      \caption{ToyFigure\cite{breaking_bad}}
      \label{fig:sfig1}
    \end{subfigure}%
    \begin{subfigure}{.125\textwidth}
      \centering
      \includegraphics[width=.9\linewidth]{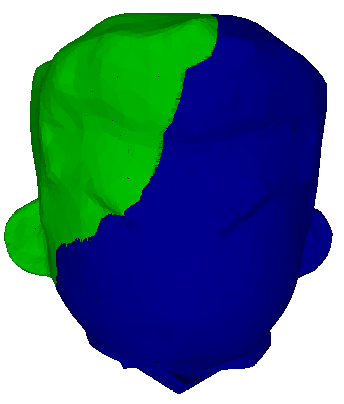}
      \caption{ToyFigure\cite{breaking_bad}}
      \label{fig:sfig1}
    \end{subfigure}%
\caption{\footnotesize A qualitative overview of our results. On the left we show reassembly of real scanned objects: (a-b) show fresco fragments from the RePAIR project\textsuperscript{2} and (c-d) show the scanned brick from the TU Wien dataset \cite{Huang2006}. On the right we shows reassembly of synthetic objects from \cite{breaking_bad}. Note that (a-d) are parts of the same object and (g-h) complete the toy figure if assembled together, acting as a starting point towards a multi-part reconstruction.}
    \label{fig:qualitative_results}
    \vspace{-0.3cm}
\end{figure*}

Despite other approaches for assembling 3D broken objects \cite{Huang2006, Papaioannou2017, altantsetseg2014pairwise} exists, we do not report a comparison with them for two reasons: \emph{i)} these algorithms \sa{have a high dependence on} particular \lp{characteristics} of the broken objects, and \emph{ii)} \sa{they are} complex to reproduce due to a large number of parameters.
Moreover, they are not suitable for \sa{assembling} synthetic objects, as they differentiate broken and intact regions of the objects based on the surface roughness \cite{Huang2006} or use feature curves to complete the reassembly \cite{Papaioannou2017}.

Although Neural Shape Mating (NSM)\cite{chen2022neural} reported promising results in the pairwise assembly task, we choose DGL as our competitor since we consider our work as a building block for the multi-part assembly task.
Moreover, NSM is using an adversarial shape loss, which \lp{requires the complete object reconstruction after pairwise assembly, while our approach, as visible in Figure \ref{fig:qualitative_results}, correctly assembles incomplete broken parts with no need for the complete object reconstruction, an important step towards real-cases multi-part assembly}.

We followed \cite{breaking_bad} for the choice of the metric using the root mean square error of the translation and the relative rotation.

\begin{table}[!ht]
\centering
\resizebox{\linewidth}{!}{
\begin{tabular}{{c}|{r}{r}{r}|{r}{r}{r}}
\toprule
& \multicolumn{3}{c}{\textbf{Relative RMSE (R)}} & \multicolumn{3}{c}{\textbf{RMSE (T)}} \\
\cmidrule(rl){2-4}\cmidrule(rl){5-7}
Category & ICP\cite{Besl1992AMF} & DGL$^\spadesuit$\cite{HuangZhan2020PartAssembly} & ours & ICP\cite{Besl1992AMF} & DGL$^\spadesuit$\cite{HuangZhan2020PartAssembly}  & ours \\
\midrule
BeerBottle & 57.028 & 78.933 & \textbf{1.62} & 1.104 & 0.073  & \textbf{0.02} \\
WineBottle & 54.262 &  84.699 & \textbf{1.58} & 0.743 & 0.024 & \textbf{0.02} \\
DrinkBottle & 60.253 & 70.014 & \textbf{1.89} & 1.288 & \textbf{0.008} & 0.033 \\
Bottle & 68.125 & 76.802 & \textbf{1.983} & 1.198 & 0.078 & \textbf{0.077} \\
Mug & 5.041 & 86.221  & \textbf{1.12} & 0.364 & 0.164 & \textbf{0.025} \\
Cookie & 12.594 &  85.707 & \textbf{1.96} & 0.632 & 0.159 & \textbf{0.043} \\
Mirror & 0.593 & 81.454  & \textbf{0.111} & 0.503 & 0.125 & \textbf{0.001} \\
ToyFigure & 208.333 & 87.972 & \textbf{1.98} & 4.123 & 0.159 & \textbf{0.079} \\
Statue & 105.582 & 89.605  & \textbf{0.66} & 2.159 & 0.149 & \textbf{0.003} \\
Vase & 30.756 & 82.218 & \textbf{0.592} & 1.496 & 0.109 & \textbf{0.002} \\
\midrule
Brick$^\clubsuit$ \cite{Huang2006} & 11.577 & 62.820 & \textbf{3.064} & 2.356 & 1.684 & \textbf{0.626} \\
{Repair}$^\clubsuit$\footnotemark[2] & 7.911 & 87.491 & \textbf{3.466} & 2.525 & \textbf{0.076} & 0.695 \\
\bottomrule
\end{tabular}}
\caption{\footnotesize Preliminary quantitative evaluation. The top rows refer to the synthetic breaking bad dataset \cite{breaking_bad} and the last two rows refer to real scanned objects. $^\spadesuit$For DGL, we take the best value for each category.  $^\clubsuit$Scanned objects, where the solution is obtained from manual alignment (Brick from TU Wien Dataset \cite{Huang2006} and fresco fragments from the RePAIR Project).}
\label{tab:bbad}
\end{table}

\sa{Quantitative results are presented in Table \ref{tab:bbad}.}
To ensure a fair comparison we list the best outcome of DGL across any pair belonging to a certain category.
\sa{Our method significantly outperforms DGL and ICP in all datasets in terms of relative rotation error and in the majority of datasets in terms of translation error.}
We note here that for some categories (Mirror, Cup, Repair) the ICP results show very low error. This happens because the broken parts are registered and completely overlap, but the solution is not satisfactory (See Figure \ref{fig:results}.e).
Additionally, we report qualitative results in Figure \ref{fig:qualitative_results} showing that our method correctly reassembles the broken parts of real and synthetic broken objects.

It is worth noting that our method is able to estimate the correct rotation and translation to assemble pairs of fragments across different datasets, while the other approaches fail in most of the cases.

\section{Conclusions}

We presented a robust method for the pairwise assembly of 3D broken objects which performs well across different datasets of both real and synthetic models.

The objective of this analysis is not to discuss which algorithm works better in which case, but rather to analyze the current situation. We note that: \textit{(i)} using an off-the-shelf approach like ICP without processing the point cloud is not a viable solution, \textit{(ii)} it is confirmed that the DGL method, which was the best performer for the published benchmark \cite{breaking_bad}, although performing well for semantic assembly, does not work for the geometric reassembly of broken objects and \textit{(iii)} using a more principled geometrical approach is a safe way to assemble broken objects.

Concerning the limitations, the proposed pipeline is sensitive to the choice of the parameters. In our experiments, we used a different set of parameters for the synthetic objects and for the real ones. There is a margin for improvements in the robustness at different steps of the pipeline.

The proposed method is presented as a building block for reassembling objects broken into multiple parts.

Extending the reassembly task to multiple broken parts following a greedy approach is under exploration. Future works include detecting non-matching surfaces and designing more principled ways of selecting the best registration among many pairs of broken objects.

\paragraph{Acknowledgements:} This work is part of a project that has received funding from the European Union’s Horizon 2020 research and innovation programme under grant agreement No 964854.
{\small
\bibliographystyle{ieee_fullname}
\bibliography{egbib}
}

\end{document}